\pdfoutput=1

\documentclass[11pt]{article}

\usepackage[final]{acl}

\usepackage{times}
\usepackage{latexsym}

\usepackage[T1]{fontenc}

\usepackage[utf8]{inputenc}

\usepackage{microtype}

\usepackage{inconsolata}

\usepackage{graphicx}
\usepackage{subcaption}

\usepackage{booktabs}

\usepackage{colortbl}

\title{How Stylistic Similarity Shapes Preferences \\in Dialogue Dataset with User and Third Party Evaluations}

\author{
 \textbf{Ikumi Numaya\textsuperscript{1}},
 \textbf{Shoji Moriya\textsuperscript{1}},
 \textbf{Shiki Sato\textsuperscript{1,2}},
 \textbf{Reina Akama\textsuperscript{1,3,4}},
 \textbf{Jun Suzuki\textsuperscript{1,4}}
\\
\\
 \textsuperscript{1}Tohoku University,
 \textsuperscript{2}CyberAgent,
 \textsuperscript{3}NINJAL,
 \textsuperscript{4}RIKEN
\\
 \small{\texttt{\{numaya.ikumi.t4, shoji.moriya.q7\}@dc.tohoku.ac.jp},} \\
   \small{\texttt{sato\_shiki@cyberagent.co.jp}}, \small{\{\texttt{akama, jun.suzuki\}@tohoku.ac.jp}}
}

\begin{document}
\maketitle
\begin{abstract}

Recent advancements in dialogue generation have broadened the scope of human–bot interactions, enabling not only contextually appropriate responses but also the analysis of human affect and sensitivity.
While prior work has suggested that stylistic similarity between user and system may enhance user impressions, the distinction between subjective and objective similarity is often overlooked.
To investigate this issue, we introduce a novel dataset that includes users’ preferences, subjective stylistic similarity based on users’ own perceptions, and objective stylistic similarity annotated by third party evaluators in open-domain dialogue settings.
Analysis using the constructed dataset reveals a strong positive correlation between subjective stylistic similarity and user preference.
Furthermore, our analysis suggests an important finding: users’ subjective stylistic similarity differs from third party objective similarity.
This underscores the importance of distinguishing between subjective and objective evaluations and understanding the distinct aspects each captures when analyzing the relationship between stylistic similarity and user preferences.
The dataset presented in this paper is available online.\footnote{\url{https://github.com/ikuminumaya/duo-dataset}}

\end{abstract}

\section{Introduction}
\label{sec:intro}

Recent advances in dialogue generation have markedly improved the ability of systems to produce appropriate and natural responses~\citep{openai2024gpt4technicalreport,shuster2022blenderbot,zhang-etal-2020-dialogpt}.
Building on this progress, personalization has become a central focus in dialogue system research, wherein system outputs are tailored to individual user preferences~\citep{tsuta-etal-2023-rethinking,du2024perltqa,chen-etal-2024-recent,firdaus-etal-2021-seprg}.
Unlike standardized evaluations such as appropriateness, individual user preferences are inherently difficult to generalize, highlighting the need for system design that accounts for user-specific characteristics.
\begin{figure}
    \centering
    \includegraphics[width=\linewidth]{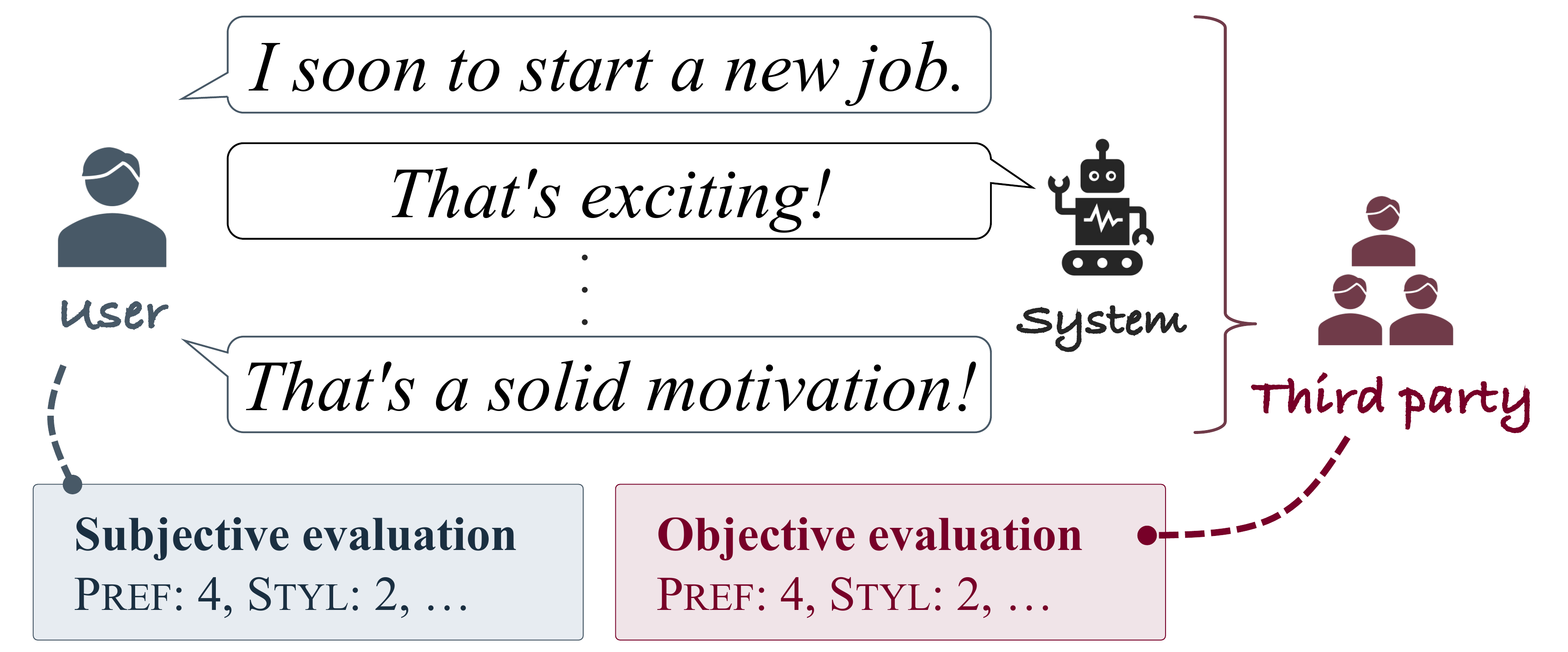}
    \caption{Overview of constructing our  dialogue dataset with users’ subjective evaluations and objective evaluations averaged over three third party annotators.}
    \label{fig:dataset-construction}
\end{figure}

Prior studies suggest that, beyond semantic content, stylistic similarity--the extent to which a system’s utterances resemble a user’s speaking style--may contribute to more positive impressions toward dialogue systems~\citep{kanezaki2024, nenkova-etal-2008-high}.
However, a critical question is often overlooked : whose perception of stylistic similarity matters, and how it affects users’ impressions.
While prior studies often rely on third party annotators to judge stylistic similarity and user engagement, it remains unclear whether users’ own perceptions of stylistic similarity or third party evaluations of it accurately capture users’ preferences. 

To investigate this question, we constructed a novel dataset incorporating users’ preferences, subjective stylistic similarity based on users’ perception, and objective ratings provided by third party annotators.
An overview of the dataset construction is shown in Figure~\ref{fig:dataset-construction}.
The dataset was based on two widely studied open-domain settings: EmpatheticDialogues~\citep{rashkin-etal-2019-towards}, focusing on empathetic communication, and Wizard of Wikipedia~\citep{dinan2018wizard}, involving knowledge-grounded conversation.
In open-domain dialogues, the primary goal is not explicit task completion but the cultivation of positive user impressions through conversational engagement~\citep{yamashita-etal-2023-realpersonachat,qian-etal-2023-harnessing,li-etal-2024-knowledge,zhang2018personalizing}.
Focusing on this setting, we use the newly constructed dataset to analyze the factors that influence users’ subjective preferences.

Correlations analysis of the subjective evaluations revealed a strong positive relationship between subjective stylistic similarity and user dialogue preference, indicating that stylistic alignment plays a key role in shaping positive user impressions.
In contrast, the correlation between objective stylistic similarity and user dialogue preference was weak, and no clear relationship emerged between subjective and objective stylistic similarity.
These findings underscore the importance of distinguishing between subjective and objective stylistic similarity in analyzing dialogue quality and user preferences.

The key contributions of this study are:
\begin{itemize}
\item We develop a multi-turn, open-domain dialogue dataset that includes both subjective evaluations by users and objective evaluations by third party annotators, covering dialogue preference and stylistic similarity.
\item We empirically demonstrate a strong association between user-perceived stylistic similarity and users’ subjective preferences.
\item We identify a discrepancy between subjective and objective stylistic similarity, emphasizing the need to consider these perspectives separately in dialogue evaluation.
\end{itemize}

\begin{figure*}
    \centering
    \includegraphics[width=0.95\linewidth]{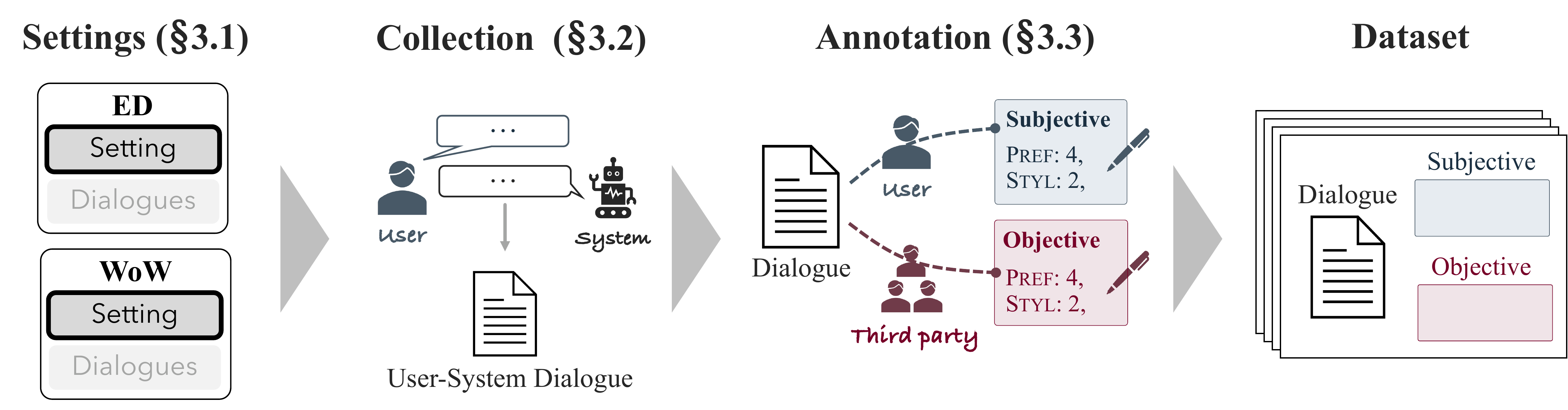}
    \caption{Overview of the entire dataset construction process.}
    \label{fig:dataset-construction-flow}
\end{figure*}
\section{Related Work}
\label{sec:relatedwork}
\paragraph{Open-domain dialogue.}

Advancements in response generation have greatly enhanced the capacity of dialogue systems to engage in coherent, contextually appropriate multi-turn interactions~\citep{openai2024gpt4technicalreport,shuster2022blenderbot,zhang-etal-2020-dialogpt}.
In open-domain dialogue research, the primary objective is not task completion but the facilitation of user-centered, engaging communication.
The Blended Skill Talk (BST) framework~\citep{smith-etal-2020-put} was introduced to evaluate dialogue systems that can blend multiple conversational skills in open-domain settings.
The framework includes personalized dialogue based on profiles~\citep{zhang2018personalizing,yamashita-etal-2023-realpersonachat,10022973}, emotion-aware responses that reflect users’ affective states~\citep{rashkin-etal-2019-towards,qian-etal-2023-harnessing,wang-etal-2025-sibyl}, and knowledge-grounded dialogue designed to support in-depth topic exploration~\citep{dinan2018wizard,li-etal-2024-knowledge,zhang-yongmei-2024-enhancing}.
These settings are widely used in multi-turn response generation studies.
In such interactions, systems are implicitly expected to maintain appropriate communication and foster user preferences.

\paragraph{Effects of entrainment on speakers.}

Entrainment, which is also referred to as accommodation or convergence, describes the gradual alignment of linguistic and prosodic features (e.g., lexical choices, response timing, and vocal pitch) between interlocutors during dialogue. 
Within the framework of Communication Accommodation Theory (CAT), \citet{giles2007communication} presented concrete examples of entrainment in human–human interaction, emphasizing its positive social effects.
Prior research has shown that entrainment can increase perceived attractiveness~\citep{Street1983TheIO}, foster interpersonal involvement~\citep{lafrance1979nonverbal}, and strengthen relational bonds~\citep{imamura2011japanese}.
Beyond these social outcomes, \citet{nenkova-etal-2008-high} introduced a method for quantifying entrainment in dialogue by analyzing word frequency patterns, demonstrating its role in enhancing perceived naturalness and task success in human dialogue.
Building on these findings, recent research has investigated the incorporation of entrainment into dialogue systems to enhance user experience~\citep{kanezaki2024}.
Aligned with this line of research, our study explores how entrainment between users and systems influences users’ subjective evaluations.

\paragraph{Style definitions.}

Text style refers to the manner in which information is conveyed beyond its literal content~\citep{ostheimer-etal-2024-text,kawano-etal-2023-analysis}.
Prior research has examined a wide range of stylistic dimensions~\citep{lai-etal-2024-style}, including formality (formal vs. informal)~\citep{rao-tetreault-2018-dear,liu-etal-2022-semi}, toxicity (toxic vs. nontoxic)~\citep{logacheva-etal-2022-paradetox,pour-etal-2023-count}, political ideology (liberal vs. conservative)~\citep{voigt-etal-2018-rtgender}, politeness (impolite vs. polite)~\citep{madaan-etal-2020-politeness,mukherjee-etal-2023-polite}, authorship style (e.g., Early Modern English vs. contemporary English)~\citep{xu-etal-2012-paraphrasing,dinu-uban-2023-computational}, and sentiment (positive vs. negative)~\citep{shen2017style}. 

\paragraph{Methods for quantifying stylistic entrainment.}

As mentioned above, entrainment in dialogue may influence impressions of the partner. In the case of stylistic entrainment, the diversity of relevant features makes consistent and automatic evaluation important.
Among the various aspects of stylistic entrainment, \citet{nasir2019modeling} proposed Local Interpersonal Distance (LID), a metric for quantifying lexical entrainment--specifically, the similarity in word choice and conveyed meaning between interlocutors.
LID employs Word Mover’s Distance (WMD)~\citep{pmlr-v37-kusnerb15,mir-etal-2019-evaluating,song-etal-2021-sentsim} to measure the distance between word embeddings of participant utterances.
Building on work, \citet{kanezaki2024} extended LID by incorporating BERTScore, allowing for more semantically sensitive quantification of lexical alignment. 
In a related effort, style-sensitive word vectors have been proposed to capture lexical stylistic features~\citep{akama-etal-2018-unsupervised}, and sentence-level embeddings have also been explored by fine-tuning BERT~\citep{zenimoto-etal-2023-style}.
Moreover, LLM-based evaluation methods, such as LLM-Eval~\citep{lin2023llm}, have recently gained traction as a mainstream approach for assessing dialogue aspects like stylistic similarity.

\begin{table}[t]
  \centering
  \small
  \setlength{\tabcolsep}{3pt} %
  \begin{tabular}{p{1.4cm} p{5.6cm}} %
    \toprule
    \textbf{Setting} & EmpatheticDialogues \\
    \midrule
    \textbf{Emotion} & Excited \\
    \textbf{Episode} &  I'm moving out west, I'm excited about it. \\
    \midrule
    \textbf{Dialogue} & \textbf{User}: \textit{I'm going to be making a big move in 2 weeks.} \\
    & \cellcolor[gray]{0.94} \textbf{System}: \textit{That's exciting! How are you feeling about it?} \\
    & \textbf{...} \\
    \midrule
    \textbf{Ratings} & \textsc{Pref.}$_\mathrm{sb}$: 1,~~~~ \textsc{Cons.}$_\mathrm{sb}$: 5,\\
    &\textsc{Styl.}$_\mathrm{sb}$: 1,~~~ \textsc{Emp.}$_\mathrm{sb}$: 4 \\
    \cmidrule{2-2}
    & \textsc{Pref.}$_\mathrm{ob}$: 3.67,~~~ \textsc{Cons.}$_\mathrm{ob}$: 4.33, ~~~\\  &\textsc{Styl.}$_\mathrm{ob}$: 3.67,~~~ \textsc{Emp.}$_\mathrm{ob}$: 4.00 \\
    \bottomrule
  \end{tabular}
  \caption{An example dialogue segment from the EmpatheticDialogues setting.
The user selects an emotion label and writes a short personal episode which serves as the context for starting the dialogue. In the dataset, the ratings include both subjective and objective evaluations.}
  \label{tab:dialog_example_ed}
\end{table}

\section{Dataset Construction}
\label{sec:dataset}

This study introduces an open-domain dialogue dataset designed to support subjective and objective evaluations.
To facilitate the analysis of the relationship between dialogue preference and stylistic similarity, data were collected under two open-domain settings, followed by annotation from both users and third party evaluators.
We named this dataset DUO (\underline{D}ialogue dataset with \underline{U}ser subjective and \underline{O}bjective evaluations).
Figure~\ref{fig:dataset-construction-flow} summarizes the overall dataset construction process described in the remainder of this section.

\subsection{Dialogue Settings}
\label{sec:setting}

We adopted two widely used open-domain dialogue settings from the Blended Skill Talk (BST) framework~\citep{smith-etal-2020-put}: EmpatheticDialogues~\citep{rashkin-etal-2019-towards} and Wizard of Wikipedia~\citep{dinan2018wizard}.
These complementary settings enable the examination of stylistic entrainment across both emotional and informational dialogue types. 
Representative examples of the collected dialogues for each setting are shown in Tables~\ref{tab:dialog_example_ed} and ~\ref{tab:dialog_example_wow}.

\paragraph{EmpatheticDialogues (ED).}

In the ED setting, the worker assumes the role of the speaker, sharing a personal emotional experience, while the dialogue system acts as the role of the empathetic listener.
Workers are presented with five randomly selected emotion labels from the set provided in the original ED framework~\citep{rashkin-etal-2019-towards}, and asked to choose one (e.g., Excited in Table~\ref{tab:dialog_example_ed}).
They are then instructed to briefly describe an emotional experience (1–3 sentences) that corresponds to the selected emotion (shown under Episode). 
The dialogue begins with the worker's utterance describing the emotional experience, followed by system responses designed to demonstrate understanding and emotional resonance.
We use this setup to assess whether the system effectively captures the user’s emotional state and responds with appropriate empathy.

\begin{table}[t]
  \centering
  \small
  \begin{tabular}{p{1.2cm} p{5.6cm}} %
    \toprule
    \textbf{Setting} & Wizard of Wikipedia \\
    \midrule
    \textbf{Topic} & Piano \\
    \midrule
    \textbf{Dialogue} & \cellcolor[gray]{0.94} \textbf{System}: \textit{The piano, invented by Bartolomeo Cristofori around 1700, is a fascinating instrument where the strings are struck by hammers controlled by a keyboard.} \\
    & \textbf{User}: \textit{Was there anything similar to the piano before this time?} \\
    & \textbf{...} \\
    \midrule
    \textbf{Ratings} & \textsc{Pref.}$_\mathrm{sb}$: 4,~~~\textsc{Cons.}$_\mathrm{sb}$: 4,\\  &\textsc{Styl.}$_\mathrm{sb}$: 4,~~~\textsc{Engag.}$_\mathrm{sb}$: 3 \\
    \cmidrule{2-2}
    & \textsc{Pref.}$_\mathrm{ob}$: 4.00,~~~ \textsc{Cons.}$_\mathrm{ob}$: 5.00,~~~\\  &\textsc{Styl.}$_\mathrm{ob}$: 2.67,~~~ \textsc{Engag.}$_\mathrm{ob}$: 4.00 \\
    \bottomrule
  \end{tabular}
  \caption{An example dialogue segment from the Wizard of Wikipedia setting.  
  Before the dialogue begins, a topic is presented to both participants. The system initiates the dialogue based on this given topic. In the dataset, the ratings include both subjective and objective evaluations.}
  \label{tab:dialog_example_wow}
\end{table}

\paragraph{Wizard of Wikipedia (WoW).}

In the WoW setting, the worker takes on the role of an apprentice, posting questions about a randomly assigned topic (e.g., Piano in Table~\ref{tab:dialog_example_wow}), while the dialogue system acts as the knowledgeable wizard.
The system always initiates the dialogue, drawing on background knowledge from the Multi-Source Wizard of Wikipedia dataset~\citep{li-etal-2024-knowledge}, which includes information from Wikipedia, Wikidata~\citep{42240}, Semantic Frames, and OPIEC~\citep{gashteovski2019opiecopeninformationextraction}.
Workers are instructed to explore the assigned topic in depth by asking questions or sharing their thoughts through the dialogue.
We use this to evaluate the system’s ability to provide informative, grounded responses and to assess its impact on user satisfaction.

\subsection{Collecting Dialogues}
\paragraph{Recruitment of participants.}
We recruited a total of 39 unique crowd workers through Amazon Mechanical Turk.\footnote{\url{https://www.mturk.com/}}
Participants were limited to individuals from the United States, Canada, or the United Kingdom, where English is the primary language.
To ensure high-quality dialogue, only workers who achieved high scores in a preliminary annotation task were invited to participate in the main study.
All workers participated anonymously, provided informed consent, and were notified that their responses might be publicly released, with all personal information fully protected.

\paragraph{Dialogue systems.}

We employed two dialogue models: GPT-4o\footnote{\url{https://platform.openai.com/docs/overview}} and Llama-3.1-70B-Instruct~\citep{grattafiori2024llama}.
GPT-4o was chosen as a representative high-performing proprietary model, while Llama-3.1-70B-Instruct was selected for its strong performance as a state-of-the-art open-weight model\citep{suresh-etal-2025-diasynth}.
Prompts were designed in accordance with prior studies~\citep{qian-etal-2023-harnessing,li-etal-2024-knowledge} and tailored to each dialogue setting.
To explore stylistic alignment, three style control conditions were introduced:
(1) instructing the system to match the user’s speaking style, 
(2) instructing it to adopt a style different from the user’s, and
(3) providing no stylistic instruction.
The influence of these conditions on stylistic similarity evaluations is demonstrated in Appendix~\ref{sec:style_control}.
The combination of the two dialogue settings, two models, and three style conditions resulted in 12 experimental configurations, enabling the collection of dialogues exhibiting a diverse range of entrainment phenomena.
For the ED setting, we employed a few-shot prompting, a method shown to yield strong performance in dialogue systems.
For the WoW setting, we employed a zero-shot prompting, which is optimized to generate knowledge-grounded responses. \footnote{Other details on the prompts are provided in Appendix~\ref{sec:Prompts}.}

\paragraph{Dialogue collection.}

Based on the dialogue settings and systems described above, we collected human-bot dialogues through interactions with crowd workers.
In principle, to minimize potential bias in evaluations, we restricted each worker to participating only once in each of the 12 experimental conditions.

\subsection{Annotation}

\paragraph{Subjective and objective stylistic similarity.}

While previous researches employed third party annotators to assess stylistic similarity and engagement, it remains unclear whether users’ own perceptions of stylistic similarity or third party evaluations of it accurately reflect users’ preferences. 
To clearly distinguish between these approaches, this study defines stylistic similarity based on who performs the evaluation.
We refer to \emph{subjective} stylistic similarity as the degree of stylistic closeness perceived by the user who actually participated in the dialogue.
\emph{Objective} stylistic similarity refers to similarity as evaluated by a third party human annotator who did not take part in the dialogue.

\begin{table}[t]
  \centering
  \small
  \begin{tabular}{lccc}
    \toprule
    & \textbf{ED} & \textbf{WoW} \\
    \midrule
    No. of dialogues        %
    & 157             & 157             \\
    No. of objective evaluations
    &50&46\\
    No. of participants     %
    & 34              & 34              \\
    No. of Unique Topics    %
    & --              & 133             \\
    No. of Unique Emotions  %
    & 30              & --              \\
    Dialogue length (min--max)         %
    & 20--23& 20--23 \\
    Avg. of dialogue length         %
    & 20.05 & 21.03 \\
    Language                %
    & English         & English         \\
    \bottomrule
  \end{tabular}
  \caption{Statistics of the constructed dataset.}
  \label{tab:dataset_stats}
\end{table}

\paragraph{Annotating by user.}
Following each dialogue, workers completed a \emph{subjective evaluation} questionnaire assessing three aspects: Preference (\textsc{Pref.}), Consistency (\textsc{Cons.}), and Stylistic similarity (\textsc{Styl.}).
For condition-specific assessments, additional labels were included: Empathy (\textsc{Emp.}) for ED, and Engagingness (\textsc{Engag.}) for WoW.
All items were rated on a five-point Likert scale, with higher scores indicating more positive evaluations.\footnote{Evaluation questions are provided in Appendix~\ref{sec:quetsion}.}
To ensure data quality, attention check questions with verifiable answers were incorporated into the questionnaire. 
Responses from workers who failed these checks were excluded from the final dataset.

\paragraph{Annotating by third party.}

In addition to the subjective evaluations collected from workers after each dialogue, we also gathered \emph{objective evaluations} from independent third party annotators who did not participate in the dialogues.
For each stylistic similarity score (1–5) from the subjective evaluation, 20 dialogues (10 from each of the WoW and ED settings) were randomly selected and independently annotated by three different workers. The final dialogue-level score was calculated by averaging the annotations from the three annotators.
To enable a direct comparison between subjective and objective evaluations, the annotation procedure (e.g., evaluation criteria and question items) followed the original subjective evaluation protocol. Annotators were given a description of the dialogue setting and were instructed to assess the dialogue as if they were the users.

\begin{table}[t]
  \centering
    \tabcolsep 2.2pt
  \small
  \begin{tabular}{lcccc}
    \toprule
    & \multicolumn{2}{c}{\textbf{Subjective}} & \multicolumn{2}{c}{\textbf{Objective}} \\
    \cmidrule(l){2-3}
    \cmidrule(l){4-5}
    \textbf{Label} & \textbf{ED} & \textbf{WoW} & \textbf{ED} & \textbf{WoW} \\
    \midrule
    \textsc{Pref.}   & 3.47 ± 1.31 & 3.96 ± 1.11 & 3.50 ± 0.79 & 3.56 ± 0.72 \\
    \textsc{Cons.}   & 4.40 ± 0.80 & 4.57 ± 0.68 & 4.55 ± 0.42 & 4.49 ± 0.57 \\
    \textsc{Styl.}   & 3.32 ± 1.31 & 3.86 ± 1.06 & 3.51 ± 0.54 & 3.18 ± 0.73 \\
    \textsc{Emp.}    & 3.87 ± 1.19 & --          & 4.10 ± 0.58 & --          \\
    \textsc{Engag.}  & --          & 3.87 ± 1.18 & --          & 3.62 ± 0.75 \\
    \bottomrule
  \end{tabular}
  \caption{Mean and standard deviation of subjective and objective evaluation scores in the ED and WoW settings.}
  \label{tab:mean_eval_by_setting}
\end{table}

\begin{table*}[t]
  \centering
  \small
  \begin{tabular}{lcccccccc}
    \toprule
     & \multicolumn{4}{c}{\textbf{ED}} & \multicolumn{4}{c}{\textbf{WoW}} \\
    \cmidrule(l){2-5}
    \cmidrule(l){6-9}
    \textbf{Label}& \textsc{Pref.}$_\mathrm{sb}$ & \textsc{Cons.}$_\mathrm{sb}$ & \textsc{Styl.}$_\mathrm{sb}$ & \textsc{Emp.}$_\mathrm{sb}$
    & \textsc{Pref.}$_\mathrm{sb}$ & \textsc{Cons.}$_\mathrm{sb}$ & \textsc{Styl.}$_\mathrm{sb}$ & \textsc{Engag.}$_\mathrm{sb}$ \\
    \midrule
    \textsc{Pref.}$_\mathrm{sb}$  & 1.00 & 0.44 & \textbf{0.75} & 0.74
                   & 1.00 & 0.50 & \textbf{0.67} & 0.80 \\
    \textsc{Cons.}$_\mathrm{sb}$  & --   & 1.00 & 0.46 & 0.54
                   & --   & 1.00 & 0.38 & 0.44 \\
    \textsc{Styl.}$_\mathrm{sb}$  & --   & --   & 1.00 & 0.66
                   & --   & --   & 1.00 & 0.59 \\
    \textsc{Emp.$_\mathrm{sb}$/Engag.$_\mathrm{sb}$} & -- & -- & -- & 1.00
                         & -- & -- & -- & 1.00 \\
    \bottomrule
  \end{tabular}
  \caption{\label{corr-labels}Spearman correlation coefficients between subjective labels in the ED and WoW settings.  
  All correlations are statistically significant with $p < 0.001$.}
\end{table*}

\subsection{Statistics of Dataset}
\label{statistics}

Table~\ref{tab:dataset_stats} shows basic statistical information of our analysis dataset DUO.
A total of 314 dialogues were gathered, evenly distributed between the two settings: 157 from ED and 157 from WoW.
Furthermore, the number of dialogues annotated by third party evaluators is 50 from ED and 46 from WoW. 
Each dialogue consists of approximately 10 turns. In the ED setting, this results in 20 utterances. In contrast, in the WoW setting, the system is designed to initiate the dialogue, leading to one additional utterance (21 utterances in total).
The ED setting includes 30 distinct emotion labels, whereas the WoW setting covers 133 unique topics, reflecting the diversity of the collected dialogues.

The mean and standard deviation for each subjective and objective evaluation label in both dialogue settings are provided in Table~\ref{tab:mean_eval_by_setting}.
\textsc{Pref.} and \textsc{Styl.} exhibited relatively high standard deviations, indicating considerable variability in user impressions, suggesting that the dataset captures a range of user-perceived preference and stylistic similarity.
In contrast, \textsc{Cons.} demonstrated high mean scores and low variance, suggesting that the dialogues were generally perceived as coherent and contextually appropriate.
These patterns were consistently observed across both dialogue settings and for both subjective and objective evaluations.
The lower standard deviations in the objective evaluations are attributed to the averaging of scores from three independent annotators.

The inter-annotator agreement (IAA) among the three annotators was measured using Krippendorff's $\alpha$, and the results showed that the $\alpha$ values for all labels were below 0.25.\footnote{Other statistics and IAA are provided in Appendix~\ref{sec:other_stats}.}
Two factors are considered to be the cause of this low agreement. First, the evaluation metrics, such as preference and style, were highly dependent on individual sensibilities. Second, because the evaluation target was the entire dialogue, consisting of about 20 utterances, judgments were prone to differ depending on which utterances the annotators focused on.

\section{Dataset Analysis}

This section presents analyses based on evaluations collected during the dataset construction.
Hereafter, subjective (sb) and objective (ob) evaluation labels are abbreviated as follows: Preference (\textsc{Pref.}$_\mathrm{sb}$ / \textsc{Pref.}$_\mathrm{ob}$), Consistency (\textsc{Cons.}$_\mathrm{sb}$ / \textsc{Cons.}$_\mathrm{ob}$)Stylistic similarity (\textsc{Styl.}$_\mathrm{sb}$ / \textsc{Styl.}$_\mathrm{ob}$), Empathy for ED (\textsc{Emp.}$_\mathrm{sb}$ / \textsc{Emp.}$_\mathrm{ob}$), and Engagingness for WoW (\textsc{Engag.}$_\mathrm{sb}$ / \textsc{Engag.}$_\mathrm{ob}$).

\subsection{Factors Influencing User’s Preference}
\label{sec:subjective-corr}
\paragraph{Preference--stylistic similarity association.}
Table~\ref{corr-labels} presents the Spearman correlations, $r$, among the subjective evaluations in the dataset. 
All correlation coefficients were statistically significant ($p < 0.001$).
Notably, \textsc{Styl.}$_\mathrm{sb}$ exhibited strong positive correlations with \textsc{Pref.}$_\mathrm{sb}$ in both dialogue settings ($r = 0.75$ for ED; $r = 0.67$ for WoW). 
These results suggest that the degree of subjective similarity between a user’s and the system’s speaking styles influences users’ favorable impressions of the dialogue.
These empirical results strongly support one of our main claims, namely, the existence of a strong association between stylistic similarity (\textsc{Styl.}$_\mathrm{sb}$) and users' subjective preferences (\textsc{Pref.}$_\mathrm{sb}$), which is one of the main claims of this paper as mentioned at the end of Section~\ref{sec:intro}.

\begin{figure}[t]
  \centering
    \small
  \begin{minipage}[t]{0.48\columnwidth}
    \centering
    \includegraphics[width=\linewidth]{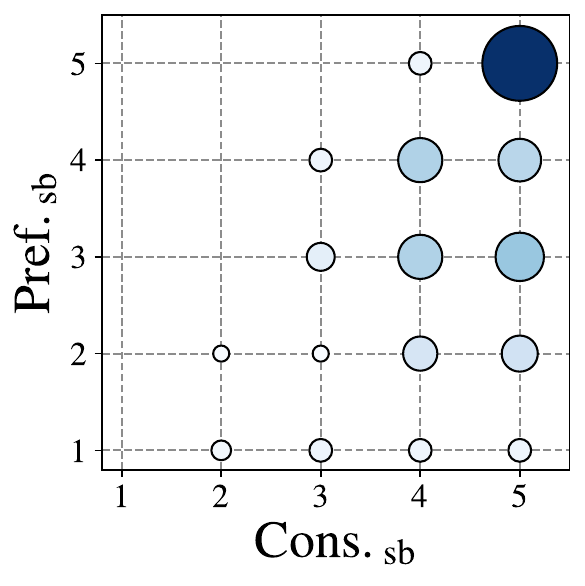}
    \par\vspace{0.3em}
    (a) ED
  \end{minipage}
  \hfill
  \begin{minipage}[t]{0.48\columnwidth}
    \centering
    \includegraphics[width=\linewidth]{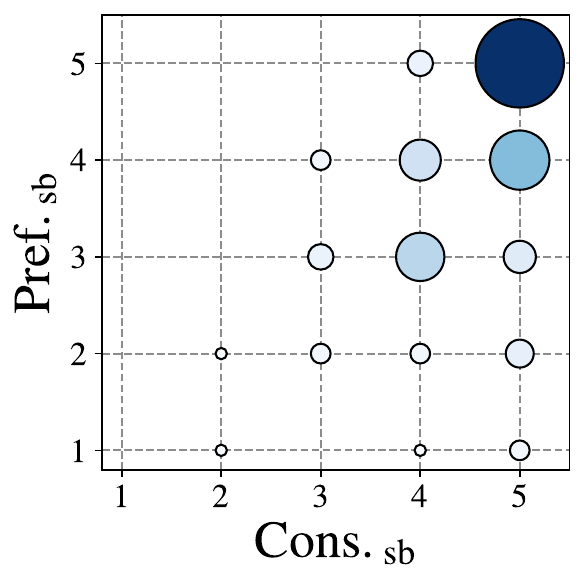}
    \par\vspace{0.3em}
    (b) WoW
  \end{minipage}

  \caption{Consistency (\textsc{Cons.}$_\mathrm{sb}$) vs. Preference (\textsc{Pref.}$_\mathrm{sb}$), where both are subjective evaluations. Larger and darker points indicate higher frequency.}
  \label{fig:consistency}
\end{figure}

\begin{figure}[t]
  \centering
    \small
  \begin{minipage}[t]{0.48\columnwidth}
    \centering
    \includegraphics[width=\linewidth]{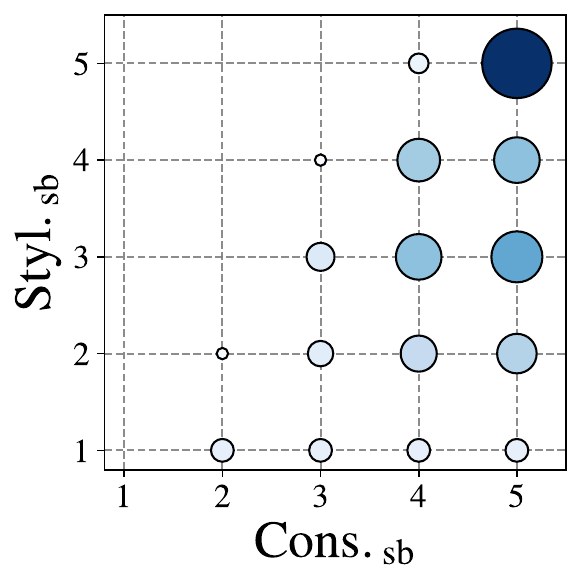}
    \par\vspace{0.3em}
    (a) ED
  \end{minipage}
  \hfill
  \begin{minipage}[t]{0.48\columnwidth}
    \centering
    \includegraphics[width=\linewidth]{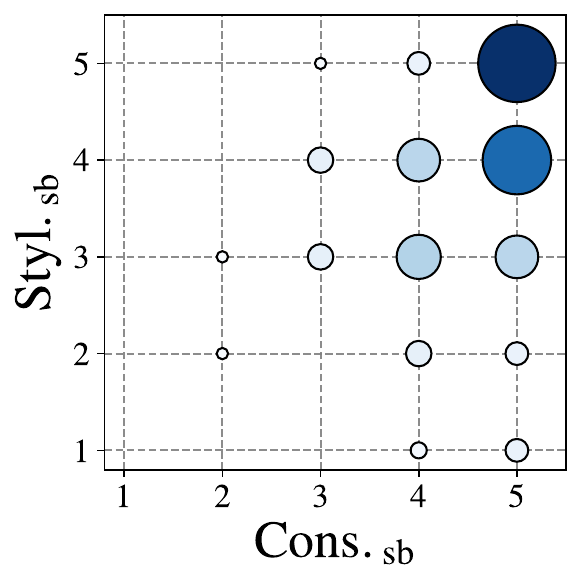}
    \par\vspace{0.3em}
    (b) WoW
  \end{minipage}

  \caption{Consistency (\textsc{Cons.}$_\mathrm{sb}$) vs. Stylistic Similarity (\textsc{Styl.}$_\mathrm{sb}$), where both are subjective evaluations. Larger and darker points indicate higher frequency.}
  \label{fig:cons-styl}
\end{figure}

\begin{table*}[t]
  \centering
  \small
  \begin{tabular}{lllllllll}
    \toprule
    & \multicolumn{4}{c}{\textbf{ED}} 
    & \multicolumn{4}{c}{\textbf{WoW}} \\
    \cmidrule(l){2-5}
    \cmidrule(l){6-9}
    \textbf{Obj. \textbackslash~ Subj.} & \textsc{Pref.}$_\mathrm{sb}$ & \textsc{Cons.}$_\mathrm{sb}$ & \textsc{Styl.}$_\mathrm{sb}$ & \textsc{Emp.}$_\mathrm{sb}$
    & \textsc{Pref.}$_\mathrm{sb}$ & \textsc{Cons.}$_\mathrm{sb}$ & \textsc{Styl.}$_\mathrm{sb}$ & \textsc{Engag.}$_\mathrm{sb}$ \\
    \midrule
    \textsc{Pref.}$_\mathrm{ob}$   & \ \ \ 0.14 & \ \ \ 0.16 & $-$0.01 & \ \ \ 0.25  & 0.35$^*$ & 0.25 & 0.28 & $0.29^{*}$ \\
    \textsc{Cons.}$_\mathrm{ob}$   & \ \ \ 0.17 & \ \ \ 0.16 & \ \ \ 0.10 & \ \ \ 0.18   & 0.47$^{***}$ & 0.21 & 0.50$^{***}$ & 0.40$^{**}$ \\
    \textsc{Styl.}$_\mathrm{ob}$   & $-$0.19 & $-$0.05 & $-$0.28 & $-$0.11 & 0.19 & 0.22 & 0.01 & 0.18 \\
    \textsc{Emp.$_\mathrm{ob}$/Engag.}$_\mathrm{ob}$ & \ \ \ 0.20 & \ \ \ 0.15 & \ \ \ 0.14 & \ \ \ 0.27 & 0.14 & 0.18 & 0.04 & 0.14 \\
    \bottomrule
  \end{tabular}
  \caption{\label{corr-objective}Spearman correlation coefficients between subjective (columns) and objective labels (rows) in the ED and WoW settings. Asterisks denote significance (* $p<.05$, ** $p<.01$, *** $p<.001$).}
\end{table*}

\paragraph{Consistency--subjective evaluations relationship.}
Table~\ref{corr-labels} shows relatively low correlations between \textsc{Cons.}$_\mathrm{sb}$ and \textsc{Pref.}$_\mathrm{sb}$ compared to other subjective evaluation pairings in both the ED and the WoW settings.
To explore this relationship, we visualized the distributions of subjective \textsc{Cons.}$_\mathrm{sb}$ and \textsc{Pref.}$_\mathrm{sb}$ in each setting, as illustrated in Figure~\ref{fig:consistency}.
Both plots reveal a consistent pattern: when \textsc{Cons.}$_\mathrm{sb}$ scores are low, \textsc{Pref.}$_\mathrm{sb}$ scores are similarly low.
In contrast, when \textsc{Cons.}$_\mathrm{sb}$ scores are high, \textsc{Pref.}$_\mathrm{sb}$ scores exhibit a broader distribution.\footnote{Other plots are provided in Appendix~\ref{sec:other_plots}.}
This pattern suggests that the consistency is a prerequisite for positive user impressions. Inconsistent dialogues tend to result in negative perceptions, regardless of other factors.
Similarly, correlations between \textsc{Cons.}$_\mathrm{sb}$ and \textsc{Styl.}$_\mathrm{sb}$ were low in Table~\ref{corr-labels}. Figure~\ref{fig:cons-styl} shows the distributions of subjective \textsc{Cons.}$_\mathrm{sb}$ and \textsc{Styl.}$_\mathrm{sb}$, both exhibiting a similar pattern, namely, lower \textsc{Cons.}$_\mathrm{sb}$ scores are associated with lower \textsc{Styl.}$_\mathrm{sb}$ scores, whereas higher \textsc{Cons.}$_\mathrm{sb}$ scores correspond to a wider spread of \textsc{Styl.}$_\mathrm{sb}$ scores.
This observation suggests that users may only perceive stylistic similarity when the system’s responses appropriately capture the dialogue context and appear natural.

\paragraph{Empathy/Engagingness--subjective evaluations.} 
The setting-specific evaluation labels, \textsc{Emp.}$_\mathrm{sb}$ for ED and \textsc{Engag.}$_\mathrm{sb}$ for WoW, also showed strong correlations with \textsc{Pref.}$_\mathrm{sb}$ ($r = 0.74$ for \textsc{Emp.}$_\mathrm{sb}$; $r = 0.80$ for \textsc{Engag.}$_\mathrm{sb}$).
Notably, the correlation coefficients are nearly equivalent to or even higher than those of the \textsc{Styl.}$_\mathrm{sb}$ (0.75 vs. 0.74 for ED and 0.67 vs. 0.80 for WoW).
These results suggest that achieving the intended communicative goals of each setting is closely linked to users’ subjective preferences.

\begin{figure}[t]
  \centering
 \small
  \begin{minipage}[t]{0.48\columnwidth}
    \centering
    \includegraphics[width=\linewidth]{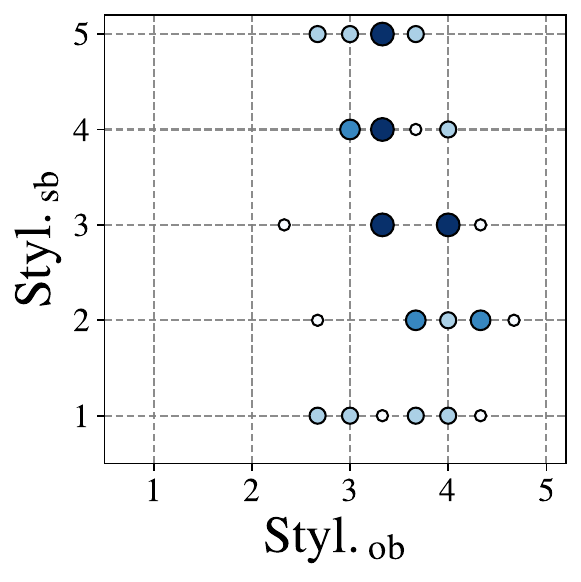}
    \par\vspace{0.3em}
    (a) ED
  \end{minipage}
  \hfill
  \begin{minipage}[t]{0.48\columnwidth}
    \centering
    \includegraphics[width=\linewidth]{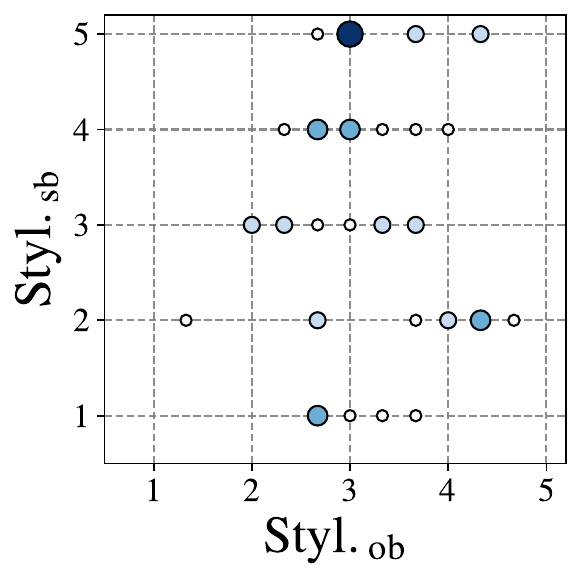}
    \par\vspace{0.3em}
    (b) WoW
  \end{minipage}

  \caption{Objective stylistic similarity (\textsc{Styl.}$_\mathrm{ob}$) vs. Subjective stylistic similarity (\textsc{Styl.}$_\mathrm{sb}$). Larger and darker points indicate higher frequency.}
  \label{fig:sims}
\end{figure}

\subsection{Gap Between Subjective and Objective}
\label{sec:gap}

Table~\ref{corr-objective} presents the Spearman correlation coefficients between subjective and objective evaluations.
This section focuses on the correlation between stylistic similarity in third party objective evaluations and subjective evaluations.
While this is a slight digression, \textsc{Cons.}$_\mathrm{ob}$ shows a positive correlation with subjective evaluations in the WoW setting, and this result suggests an interesting direction for future work.
\begin{table}[t]
  \centering
  \small
  \setlength{\tabcolsep}{3pt} %
  \begin{tabular}{p{1.4cm} p{5.6cm}} %
    \toprule
    \textbf{Dialogue} & \textbf{User}: \textit{I was prepared to do my taxes this past weekend. I had all my documents ready.} \\
    & \cellcolor[gray]{0.94} \textbf{System}: \textit{\textbf{Ah}, taxes! Something unexpected came up, \textbf{huh?}} \\
     & \textbf{User}: \textit{\textbf{No}, it was a couple days before tax day.} \\
    & \cellcolor[gray]{0.94} \textbf{System}: \textit{\textbf{Ah}, cutting it close, \textbf{huh?}} \\
    & \textbf{User}: \textit{\textbf{Oh} I don't know, I gave myself a few days buffer.} \\
    &\textbf{...}\\
    \midrule
    \textbf{Ratings} & \textsc{Styl.}$_\mathrm{sb}$: 2 \quad \textsc{Styl.}$_\mathrm{ob}$: 4.33 (4,4,5) \\
    \bottomrule
  \end{tabular}
  \caption{Example dialogue in the ED setting where both subjective stylistic similarity is low, while objective stylistic similarity is high.}
  \label{tab:example_gap}
\end{table}

\begin{table}[t]
  \centering
  \small
  \setlength{\tabcolsep}{3pt} %
  \begin{tabular}{p{1.4cm} p{5.6cm}} %
    \toprule
    \textbf{Dialogue} & \textbf{User}: \textit{I was surprised when my husband gave me a surprise birthday party to celebrate my 50th birthday.} \\
    & \cellcolor[gray]{0.94} \textbf{System}: \textit{\textbf{Wow}, that sounds amazing! It must have been a wonderful surprise to celebrate such a special milestone.} \\
     & \textbf{User}: \textit{\textbf{Yes}, I had no idea he had been planning this surprise for me and I was truly surprised!} \\
    & \cellcolor[gray]{0.94} \textbf{System}: \textit{\textbf{That's awesome!} He must really know how to make you feel special.} \\
    & \textbf{User}: \textit{\textbf{Yes}, I feel blessed to have my husband and he is truly my soulmate. I felt so surprised that he planned this special party for me.} \\
    &\textbf{...}\\
    \midrule
    \textbf{Ratings} & \textsc{Styl.}$_\mathrm{sb}$: 5 \quad \textsc{Styl.}$_\mathrm{ob}$: 2.67 (2,2,4) \\
    \bottomrule
  \end{tabular}
  \caption{Example dialogue in the ED setting where both subjective stylistic similarity is high, while objective stylistic similarity is low.}
  \label{tab:example_high_low}
\end{table}

\begin{table*}[t]
  \centering
  \small
  \begin{tabular}{lllllllll}
    \toprule
    & \multicolumn{4}{c}{\textbf{ED}} & \multicolumn{4}{c}{\textbf{WoW}} \\
    \cmidrule(l){2-5}
    \cmidrule(l){6-9}
& \textsc{Pref.} & \textsc{Cons.}& \textsc{Styl.} & \textsc{Emp.} 
    & \textsc{Pref.} & \textsc{Cons.} & \textsc{Styl.} & \textsc{Engag.} \\
    \midrule
    \textbf{Subjective}    & 0.22$^{**}$ & 0.17$^*$ & 0.17$^*$ & 0.26$^{**}$ & 0.27$^{***}$ & 0.20$^*$ & 0.17$^*$ & 0.29$^{***}$ \\
    \textbf{Objective}    & 0.23 & 0.25 & 0.07 & 0.36$^{*}$ & 0.47$^{**}$ & 0.45$^{**}$ & 0.22 & 0.38$^{**}$ \\
    \bottomrule
  \end{tabular}
  \caption{\label{corr-auto}Spearman correlation coefficients between stylistic similarity measures, {Subjective} and {Objective}, (rows) and users’ subjective evaluations (columns) in the ED and WoW settings.
Asterisks indicate significance (* $p<.05$, ** $p<.01$, *** $p<.001$).}
\end{table*}

\paragraph{Subjective vs. objective stylistic similarity.}
In Table~\ref{corr-objective}, a weak but positive correlation was observed between \textsc{Pref.}$_\mathrm{sb}$ and \textsc{Styl.}$_\mathrm{ob}$, particularly in the WoW setting ($r = 0.28$).
In contrast, the correlation between \textsc{Styl.}$_\mathrm{sb}$ and \textsc{Styl.}$_\mathrm{ob}$ remained low.
This indicates a gap between the users’ perceived degree of stylistic similarity during actual participation in the dialogue and the similarity assessed by third party annotators.
Notably, although both subjective and objective stylistic similarity are based on human evaluations, the observed discrepancy suggests they capture different perspectives.
These findings underscore the importance of distinguishing between subjective and objective evaluations when analyzing stylistic similarity, which is also one of the main claims of this paper as mentioned at the end of Section~\ref{sec:intro}.

\paragraph{Explaining the gap in stylistic similarity.}

As discussed earlier, the correlation between \textsc{Styl.}$_\mathrm{sb}$ and \textsc{Styl.}$_\mathrm{ob}$ was not positive, and the two measures themselves showed no significant relationship.
To explore this discrepancy, we conducted a qualitative analysis.
Figure~\ref{fig:sims} presents scatter plots comparing \textsc{Styl.}$_\mathrm{sb}$ and \textsc{Styl.}$_\mathrm{ob}$.
Notably, several dialogues rated low on \textsc{Styl.}$_\mathrm{sb}$ (scores of 1 or 2) still received relatively high \textsc{Styl.}$_\mathrm{ob}$ scores (above 2). This suggests that users’ perceptions of stylistic mismatch may diverge from those of third party annotators. 
Additionally, this suggests that users tend to make more polarized, intuitive judgments based on whether the system’s style feels personally similar. In contrast, third party annotators, who rely solely on dialogue text, appear to assess stylistic similarity in a more graded and continuous manner.
While this analysis highlights a gap between subjective and objective stylistic similarity, similar discrepancies may arise in other tasks due to differing evaluative perspectives, such as those between dialogue participants and external annotators.

\paragraph{Examples of the gap in stylistic similarity.}
Tables~\ref{tab:example_gap} and \ref{tab:example_high_low} show examples of dialogues that exhibit gaps between subjective and objective stylistic similarity scores. 
In Table~\ref{tab:example_gap}, the objective stylistic similarity score assigned by third party annotators is higher than the subjective rating provided by the user. 
At the beginning of the dialogue, the user adopts a formal style (e.g., \textit{`No'}, avoiding contractions), while the system responds in a more casual manner (e.g., \textit{`Ah'} or \textit{`huh'}). 
As the conversation progresses, the user shifts to a more casual tone (e.g., \textit{`Oh'}), seemingly adapting to the system’s style. 
From the user’s perspective, the initial stylistic mismatch, and the perceived need to adjust, may have contributed to a lower subjective similarity rating. 
In contrast, third party annotators may have focused on the later-stage stylistic convergence, resulting in a higher objective rating.
In Table~\ref{tab:example_high_low}, the opposite pattern is observed: the user assigns a higher stylistic similarity score than the third party annotators. 
Here, the system produces positive expressions (e.g., \textit{`Wow'}, \textit{`That’s awesome'}), and the user replies with similarly positive responses (e.g., \textit{`Yes'}). 
This emotional alignment may have led the user to perceive high stylistic similarity. 
However, third party annotators may have noticed subtle mismatches in formality, such as the difference in tone between \textit{`Wow'} and \textit{`Yes'}, and rated the similarity lower.
These examples highlight how discrepancies in evaluation can emerge depending on the rater’s perspective. 
They underscore the challenge of achieving consistent assessments and the importance of distinguishing between subjective evaluations from dialogue users and objective evaluations by third party annotators.

\subsection{Automatic Evaluation}

The primary focus of this paper is on the human evaluation setting.
Additionally, this section briefly discusses the automatic evaluation setting.

\paragraph{Evaluation method.}

We followed LLM-Eval~\citep{lin2023llm} as an automatic evaluation method.
LLM-Eval has been shown to produce dialogue quality scores that better align with objective human evaluations compared to traditional automatic metrics such as BLEU-4~\citep{papineni-etal-2002-bleu} and ROUGE-L~\citep{lin-2004-rouge}.
In this experiment, GPT-4o was used as the underlying LLM to generate LLM-Eval scores, specifically producing scores for the label ``stylistic similarity.''

\paragraph{Results and discussions.}
Table~\ref{corr-auto} shows the correlation coefficients between the quantified similarity score automatically provided by LLM-Eval and both users’ subjective and objective evaluations.
We find that LLM-Eval’s scores correlate more strongly with objective evaluations.
This is unsurprising, given that the method itself is objective rather than subjective.
These results again highlight the importance of considering human evaluations, whether obtained in subjective or objective settings.
Different settings may lead to different conclusions.
They also support our claim that there is a discrepancy between subjective and objective stylistic similarity, emphasizing the need to consider these perspectives separately in dialogue evaluation.
We also note that we have never claimed that our results conflict with previous findings of stylistic similarity correlates with human preference, as there is a certain degree of correlation in the objective settings, and prior studies primarily focus on objective stylistic similarity.

\section{Conclusion}

This study focused on stylistic similarity as a key factor influencing individual users’ dialogue preferences. While prior work has pointed to the potential relevance of stylistic similarity, it remained unclear whether users’ own perception of stylistic similarity or third party assessments play a more critical role. To explore this question, we collected human–bot dialogues in open-domain settings and obtained both users’ subjective preferences and stylistic similarity ratings, as well as third party annotations. 
Our findings revealed a strong correlation between subjective stylistic similarity and user preference, whereas objective stylistic similarity showed only a weak correlation. 
Moreover, there was no significant association between subjective and objective stylistic similarity. 
They suggest that subjective and objective evaluations reflect different aspects of style, underscoring the need to analyze them separately.

The observed discrepancies suggest that whether the evaluation is made by a participant or a third party may matter not only for stylistic similarity but also for other aspects of dialogue assessment. 
Future work could consider the impact of such perspective differences on system evaluation. %

\section*{Limitations}

In this study, the scope of the dialogues analyzed is restricted to open-domain dialogues, excluding task-oriented dialogues, such as those focused on information provision or problem-solving, which were not sufficiently addressed.
As a result, the proposed methodology and evaluation findings may not be directly applicable to task-oriented dialogue systems or other specific conversational contexts.

\section*{Ethical Considerations}

Participants were recruited via the crowdsourcing platform Amazon Mechanical Turk to collect dialogue data and evaluations.
Two tasks were conducted: the subjective evaluation task, where participants interacted with the system and rated their experience, and the objective annotation task, where third party annotators evaluated 10 dialogues per session.
A detailed task description, outlining structure and key instructions, was provided to all participants. Informed consent was obtained, including assurances that personal information would be protected and that the collected data might be made publicly available in the future. For third party annotators, additional consent was obtained to inform them that the dialogues may contain potentially sensitive content.
The subjective evaluation task and the objective annotation task were designed to take approximately 15 and 20 minutes respectively to complete, with participants compensated at an estimated rate of \$15 per hour (i.e., \$3.75 and \$5.00 per task, respectively).

The collected data may include potentially sensitive content, such as personal views on political orientation or social values. These data are solely used for analyzing evaluations of dialogue and are not employed for model training or response generation.
Any utterances containing discriminatory or overtly offensive language were manually filtered. Furthermore, all personally identifiable information, including user names and named entities related to the individual, has been masked to ensure anonymity.

We plan to release the DUO dataset in a format that allows reuse for non-commercial and non-profit research purposes only.
To promote responsible use, we will explicitly note that the dataset may contain culturally or ideologically sensitive content, and encourage users to review it carefully and exercise appropriate judgment before use.

\section*{Acknowledgments}
This research was supported by JST Moonshot R\&D Grant Number JPMJMS2011-35 (fundamental research), JSPS KAKENHI Grants JP25K21263, JST BOOST JPMJBY24A1.
\clearpage

\bibliography{custom}

\newpage

\appendix

\section{Influence of Style Control Conditions}
\label{sec:style_control}

Table~\ref{tab:mean_eval_by_system} shows the mean subjective stylistic similarity for each system. We controlled the style of the systems used for dialogue collection with the following three types of prompts:
(1) Aligned: Instructing the system to match the user's speaking style.
(2) Not Aligned: Instructing it to adopt a style different from the user's.
(3) Neutral: Providing no stylistic instruction.
The mean values indicate that these prompts successfully generated a range of subjective stylistic similarities. In the WoW setting, in particular, we observed a trend where the mean stylistic similarity increased in the intended order: Not aligned, Neutral, and then Aligned.

\begin{table}[ht]
  \centering
    \tabcolsep 2.2pt
  \small
  \begin{tabular}{lcccc}
    \toprule
    & \multicolumn{2}{c}{\textbf{ED}} & \multicolumn{2}{c}{\textbf{WoW}} \\
    \cmidrule(l){2-3}
    \cmidrule(l){4-5}
    \textbf{Prompt}& \textbf{GPT-4o} & \textbf{Llama-3.1} & \textbf{GPT-4o} & \textbf{Llama-3.1} \\
    \midrule
    \textbf{Not Aligned}   & 3.60 & 3.57 & 3.71 & 3.87 \\
    \textbf{Neutral}   & 2.61 & 3.32 & 3.74 & 3.89 \\
    \textbf{Aligned}   & 3.59 & 3.38 & 3.86 & 4.13 \\
    \bottomrule
  \end{tabular}
  \caption{Mean subjective stylistic similarity for each prompt condition (Aligned, Neutral and Not Aligned ). }
  \label{tab:mean_eval_by_system}
\end{table}

\section{Prompts}
\label{sec:Prompts}
\subsection{Stylistic Similarity Prompt}
\label{sec:prompt_style}

The dialogue systems used in this study were prompted to respond in either a similar or dissimilar speaking style relative to the user’s utterances.
Examples of the prompts are shown in Figure~\ref{fig:prompt-style-control}.
The illustrative style categories follow the classification proposed by \citet{lai-etal-2024-style}.

\subsection{EmpatheticDialogue}
\label{sec:prompt_ed}

Figure~\ref{fig:prompt-template-ed} shows the prompt used for the system in the ED setting.
This prompt is based on the template proposed by \citet{qian-etal-2023-harnessing}, which demonstrated strong performance for existing dialogue systems using a few-shot prompting approach.
The dialogue examples were randomly sampled from the training split of the ED dataset.

\subsection{Wizard of Wikipedia}
\label{sec:prompt_wow}

Figure~\ref{fig:prompt-template-wow} shows the prompt used for the system in the Wizard of Wikipedia setting.
This prompt is based on the template proposed by \citet{li-etal-2024-knowledge}, which demonstrated strong performance in generating responses grounded in factual knowledge for existing dialogue systems.
For each topic, knowledge entries were sourced from the Multi-Source Wizard of Wikipedia dataset~\citep{li-etal-2024-knowledge}, which aggregates information from multiple sources, and incorporated into the prompt.

\begin{figure}[t]
  \centering
  \fbox{%
  \begin{minipage}{0.95\linewidth}
  \small
  \ttfamily
Please make the style match from (\textbf{or} be different) user's Style includes things like the following:\\
\hspace*{2em}- Formality: informal vs formal\\
\hspace*{2em}- Politeness: impolite vs polite\\
\hspace*{2em}- Gender: masculine vs feminine\\
\hspace*{2em}- Biasedness: biased vs neutral\\
\hspace*{2em}- Toxicity: offensive vs non-offensive
  \end{minipage}
  }
  \caption{Prompt snippet for controlling stylistic alignment in system responses.}
  \label{fig:prompt-style-control}
\end{figure}

\begin{figure}[t]
  \centering
  \fbox{%
  \begin{minipage}{0.95\linewidth}
  \small
  \ttfamily
This is an empathetic dialogue task: The first worker (Speaker) is given an emotion label and writes his own description of a situation when he has felt that way. Then, Speaker tells his story in a conversation with a second worker (Listener). The emotion label and situation of Speaker are invisible to Listener. Listener should recognize and acknowledge others' feelings in a conversation as much as possible.\\

Now you play the role of Listener, please give the corresponding response according to the existing context. You only need to provide the next round of response of Listener.\\

The following is the existing dialogue context:\\

Instance 1: \\
...\\
Instance 2: \\
...\\
Instance 3: \\
... \\
Instance 4: \\
...\\
Instance 5: \\
...\\
""" \\
\{Dialog history\} \\
""" \\
""" \\
\{Stylistic similarity prompts\} \\
""" \\
You MUST keep response as short as possible.

  \end{minipage}
  }
  \caption{Prompt for the system in the ED setting.}
  \label{fig:prompt-template-ed}
\end{figure}

\begin{figure}[t]
  \centering
  \fbox{%
  \begin{minipage}{0.95\linewidth}
  \small
  \ttfamily
The following is the conversation between the “Wizard", a knowledgeable speaker who can access Wikipedia knowledge sentences to chat to with the “Apprentice", who does not have access to Wikipedia. The conversation topic is \{Topic\}.\\

This is their conversation history:\\
""" \\
\{Dialog history\} \\
""" \\

Here is some retrieved Wikipedia knowledge for the Wizard. The Wizard can choose any subset of the following knowledge. It’s also allowed to not choose any of them.\\
""" \\
\{Knowledge\} \\
""" \\
""" \\
\{Stylistic similarity prompts\} \\
""" \\
Given the knowledge above, make a very brief, such as one sentence, natural response for the Wizard. Not all information in the chosen knowledge has to be used in the response. \\

The Wizard’s response is:

  \end{minipage}
  }
  \caption{Prompt for the system in the WoW setting}
  \label{fig:prompt-template-wow}
\end{figure}

\begin{table}[t]
  \centering
  \small
  \setlength{\tabcolsep}{2pt}
  \begin{tabular}{p{2.5cm} p{4.8cm}} %
    \toprule
    \textbf{Label} & \textbf{Question}\\
    \midrule
    \textbf{Preference}&Was the dialogue preferable for you?\\
    \midrule
    \textbf{Stylistic similarity} &Was the system's speaking style similar to yours?\\
    \midrule
    \textbf{Consistency}&Was the system consistent in the information it provided throughout the dialogue?\\
    \midrule
    \textbf{Empathy} &Did the responses show understanding of the feelings of you talking about your experience?\\
    \midrule
    \textbf{Engagingness} &How engaging did you find the conversation?\\
    \bottomrule
\end{tabular}
  \caption{Questions presented to users for evaluating different aspects of the dialogue.}
  \label{tab:questionnaire}
\end{table}

\section{Post-Dialogue Questionnaire}
\label{sec:quetsion}

Table~\ref{tab:questionnaire} presents the specific question items used to collect both the subjective evaluations from dialogue participants and the objective evaluations from third party annotators for each label in the dataset. For the objective evaluations, annotators were instructed to answer the questions from the perspective of a user interacting with the system.
The question for Consistency was adapted from the FED metric~\citep{mehri-eskenazi-2020-unsupervised}, while the Empathy item was based on the original prompt used in ED~\citep{rashkin-etal-2019-towards}.
All responses were collected on a 5-point Likert scale (1: not at all, 2, 3: somewhat, 4, 5: very much).

\section{Other Statistics of Dataset}
\label{sec:other_stats}

Table~\ref{tab:other_stats} presents statistics of our dataset that were not discussed in the main text. Focusing on utterance length, utterances in the WoW setting tend to be longer. This is likely due to the nature of the task, where the Wizard provides informative responses related to the given topic.

Table \ref{tab:agreement} shows the results for the inter-annotator agreement among the three annotators for the objective evaluation, as measured by Krippendorff's $\alpha$. The evaluation criteria include Pref$_\mathrm{ob}$, Cons$_\mathrm{ob}$, Styl$_\mathrm{ob}$, Emp$_\mathrm{ob}$, Engag$_\mathrm{ob}$. The results showed that the $\alpha$ values for all labels were below 0.25.

\begin{table}[t]
  \centering
  \small
  \begin{tabular}{lcc}
    \toprule
    & \textbf{ED} & \textbf{WoW} \\
    \midrule
    No. of utterances       & 3148            & 3302            \\
    No. of tokens           & 44640           & 63157           \\
    Utterance length        & 1--76           & 1--146  \\
    Avg. of Utterance length&14.18            &19.13\\
    Vocabulary              & 3819            & 6439            \\
    Type-Token ratio        & 0.086          & 0.10          \\
    Herdan’s C              & 0.77          & 0.79          \\
    \bottomrule
  \end{tabular}
  \caption{Additional statistics of the dataset.}
  \label{tab:other_stats}
\end{table}

\begin{table}[t]
  \centering
  \setlength{\tabcolsep}{2.0pt}
  \small
  \begin{tabular}{l ccccc}
    \toprule
    & \textsc{\textbf{Pref}.}$_\mathrm{ob}$
    & \textsc{\textbf{Cons}.}$_\mathrm{ob}$
    & \textsc{\textbf{Styl}.}$_\mathrm{ob}$ 
    & \textsc{\textbf{Emp}.}$_\mathrm{ob}$
    & \textsc{\textbf{Engag}.}$_\mathrm{ob}$
    \\
    \midrule
    \textbf{ED}
    & 0.20
    & 0.15
    & $-$0.07
    & $-$0.09
    & -
    \\
    \textbf{WoW}
    & 0.11
    & 0.24
    & 0.09
    & -
    & 0.13
    \\
  \bottomrule
  \end{tabular}
  \caption{Inter-annotator agreement for the objective evaluations, measured by Krippendorff's $\alpha$.}
  \label{tab:agreement}
\end{table}

\section{Scatter Plots of Subjective Evaluations}
\label{sec:other_plots}

\begin{figure}[ht]
    \centering
    \begin{subfigure}[b]{0.45\columnwidth}
        \includegraphics[width=\linewidth]{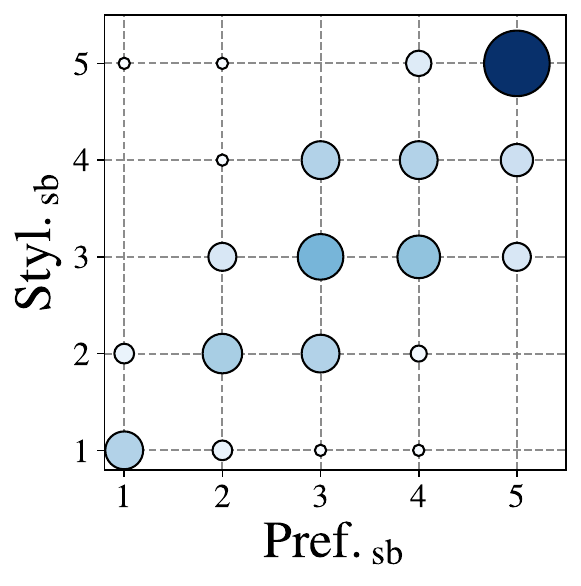}
        \caption{ED}
    \end{subfigure}
    \hfill
    \begin{subfigure}[b]{0.45\columnwidth}
        \includegraphics[width=\linewidth]{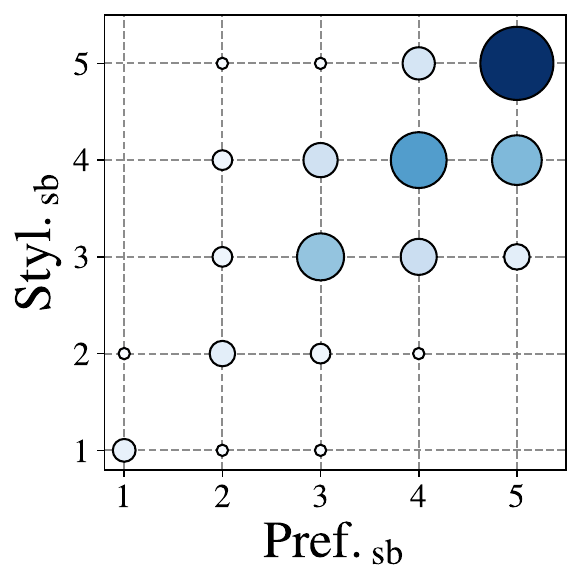}
        \caption{WoW}
    \end{subfigure}
    \vspace{-2mm}
    \caption{Subjective preference vs. stylistic similarity.}
    \label{fig:scatter_pre_styl}
\end{figure}

\begin{figure}[ht]
    \centering
    \begin{subfigure}[b]{0.45\columnwidth}
        \includegraphics[width=\linewidth]{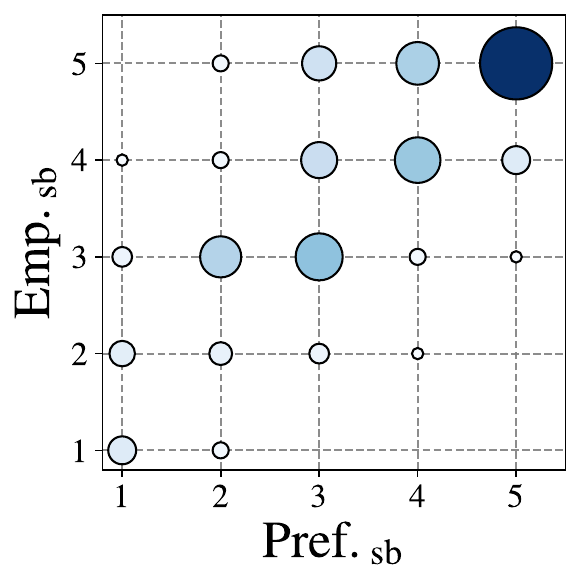}
        \caption{ED}
    \end{subfigure}
    \hfill
    \begin{subfigure}[b]{0.45\columnwidth}
        \includegraphics[width=\linewidth]{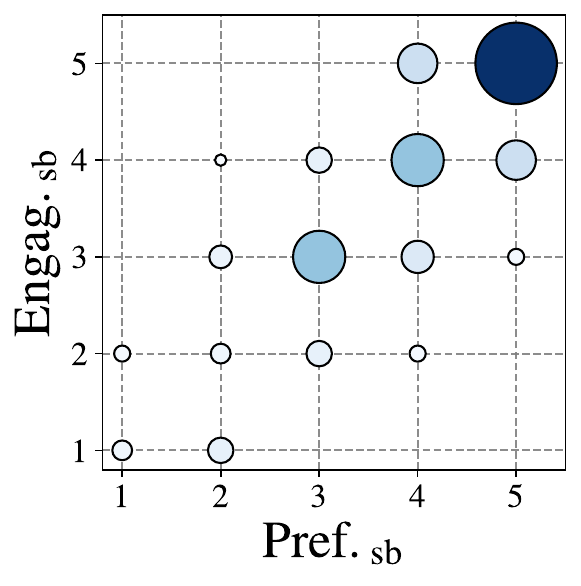}
        \caption{WoW}
    \end{subfigure}
    \vspace{-2mm}
    \caption{Subjective preference vs. empathy on ED and engagingness on WoW.}
    \label{fig:scatter_pre_quali}
\end{figure}

In Section~\ref{sec:subjective-corr}, we analyzed the correlations between subjective evaluation labels, as shown in Table~\ref{corr-labels}.
Scatter plots for label pairs not discussed in the main text are presented in Figures~\ref{fig:scatter_pre_styl} and \ref{fig:scatter_pre_quali}.

\end{document}